\documentclass[runningheads]{llncs}
\usepackage[T1]{fontenc}
\usepackage{graphicx}

\usepackage{subcaption}
\usepackage{algorithm}
\usepackage{algorithmic}
\usepackage{booktabs}
\usepackage{multirow}
\begin{document}
%
\title{Divide-and-Conquer Strategy for Large-Scale Dynamic Bayesian Network Structure Learning}
\titlerunning{Divide-and-Conquer Strategy for Large-Scale DBN Structure Learning}
%
\author{Hui Ouyang\inst{1}
\and Cheng Chen\inst{2}
\and Ke Tang\inst{1}}

\authorrunning{H. Ouyang et al.}
%
\institute{Department of Computer Science and Engineering, Southern University of Science and Technology, Shenzhen 518055, China \\
\email{ouyangh2021@mail.sustech.edu.cn,tangk3@sustech.edu.cn}
\and Research Institute of Trustworthy Autonomous Systems, Southern University of Science and Technology, Shenzhen 518055, China \\
\email{chenc3@sustech.edu.cn}
}
\maketitle              
\begin{abstract}

Dynamic Bayesian Networks (DBNs), renowned for their interpretability, have become increasingly vital in representing complex stochastic processes in various domains such as gene expression analysis, healthcare, and traffic prediction. 
Structure learning of DBNs from data is challenging, particularly for datasets with thousands of variables. 
Most current algorithms for DBN structure learning are adaptations from those used in static Bayesian Networks (BNs), and are typically focused on small-scale problems. 
In order to solve large-scale problems while taking full advantage of existing algorithms, this paper introduces a novel divide-and-conquer strategy, originally developed for static BNs, and adapts it for large-scale DBN structure learning.
In this work, we specifically concentrate on 2 Time-sliced Bayesian Networks (2-TBNs), a special class of DBNs. 
Furthermore, we leverage the prior knowledge of  2-TBNs to enhance the performance of the strategy we introduce. 
Our approach significantly improves the scalability and accuracy of 2-TBN structure learning. 
Experimental results demonstrate the effectiveness of our method, showing substantial improvements over existing algorithms in both computational efficiency and structure learning accuracy. 
On problem instances with more than 1,000 variables, our approach improves two accuracy metrics by $74.45\%$ and $110.94\%$ on average , respectively, while reducing runtime by $93.65\%$ on average.

\end{abstract}

\section{Introduction}

Amidst advancements in deep neural networks, particularly the emergence of Large Language Models (LLMs) ~\cite{openai2023gpt4,NingLiu2023}, deep learning has revealed unexpected potential. 
However, interpretability continues to challenge AI researchers~\cite{zhang2021survey,NingLiu2023a}. 
Causal models, represented by Bayesian Networks (BNs)~\cite{pearl1985bayesian}, are noted for their superior interpretability. 
Within this domain, Dynamic Bayesian Networks (DBNs)~\cite{murphy2002dynamic} stand out as a broadly applicable and adaptable model class, adept at representing complex stochastic processes. 
Their utilization spans various fields, including gene expression analysis~\cite{ajmal2021dynamic}, medical care~\cite{romero2023dynamic}, and traffic prediction~\cite{lin2023dynamic}, and etc.
Current algorithms for DBN structure learning are primarily adaptations of static BN structure learning algorithms. 
Identifying the graphical structure of BNs, especially under causal assumptions, remains a significant challenge~\cite{kitson2023survey}. 
Most of these algorithms have focused on small-scale problems.
However, real-world scenarios often demand causal structure learning for large-scale problems, as seen in applications like MRI image interpretation~\cite{ramsey2017million}, human genome analysis~\cite{bernaola2020learning}, and industrial Internet data mining~\cite{zhu2021efficient}.

Recently, a novel divide-and-conquer strategy for large-scale static BN structure learning has been proposed, termed Partition-Estimation-Fusion (PEF)~\cite{gu2020learning}.
Employing a divide-and-conquer strategy yields distinct advantages. 
When the integrated structural learning algorithm, used for solving subproblems, has a time complexity beyond linear, enhanced scalability is expected. 
Ignoring causal insufficiency, subproblems are considerably less complex compared to the original problem. 
Effectively dividing nodes to weaken inter-cluster connectivity relative to intra-cluster can reduce the impact of divide-induced causal insufficiency on overall accuracy. 
Notably, the estimation phase of the PEF, which involves solving different subproblems, is highly parallelizable.
For DBN structure learning, works like \cite{friedman1998learning,trabelsi2013dynamic} have transformed the problem into two separate static BN learning tasks. 
In this work, we assert that the PEF strategy can be effectively adapted for the structure learning of 2 Time-sliced Bayesian Networks (2-TBN), a special class of DBNs, to yield strong performance. 
Furthermore, we enhance the PEF strategy for DBN structure learning by leveraging prior knowledge of 2-TBN.
The contributions of this paper are threefold: 
\begin{enumerate}
    \item Introducing the divide-and-conquer PEF strategy from static BNs to DBNs;
    \item Utilizing prior knowledge of 2-TBN to augment the PEF strategy for DBNs;
    \item Conducting experiments to validate the effectiveness of our proposed strategy in DBN structure learning.
\end{enumerate}

The remainder of this paper is organized as follows.
Section~\ref{sec:prw} provides background on BNs, DBNs, and their structure learning challenges, along with a review of related work. 
Section~\ref{sec:pm} details our proposed divide-and-conquer strategy for 2-TBNs. 
Experimental validation of our approach is presented in Section~\ref{sec:exp}, and Section~\ref{sec:con} concludes the paper with a summary and discussion of future research directions.

\section{Preliminaries and Related Work}\label{sec:prw}

\subsection{Bayesian Network}

Bayesian Networks (BNs)~\cite{pearl1985bayesian} are a theoretical model used to describe probabilistic relationships, providing a clear representation of causal information.
For a set of $n$ random variables $X = \{X_1, X_2, \dots, X_n\}$, a BN $B = (G, \Theta)$ consists of a Directed Acyclic Graph (DAG) $G = (V, E)$ and a parameter set $\Theta$. 
Here, $V$ and $E$ respectively denote the vertex and edge sets of the DAG $G$. 
Each vertex $V_i \in V$ represents a random variable $X_i \in X$, and each edge $(V_i, V_j) \in E$ indicates a directed causal relationship between $X_i$ and $X_j$. 
The dependency of each variable $X_i$ is solely on variable(s) $X_{pa(i)} \in X$ associated with its parent nodes $V_{pa(i)} \in V$. 
The parameter set $\Theta = \{P(X_i | X_{pa(i)})\}$ encompasses all conditional probabilities of the variables given the states of their parents in $G$. 
Adhering to the first-order Markov property, these random variables enable a compact representation of the joint probability distribution over all variables:
\begin{equation}
    P(X) = \prod_{i=1}^{n} P(X_i | X_{pa(i)})
\end{equation}

\subsection{Dynamic Bayesian Network}

\begin{figure}[t]
    \centering
    \begin{subfigure}{0.38\textwidth}
        \includegraphics[width=\linewidth]{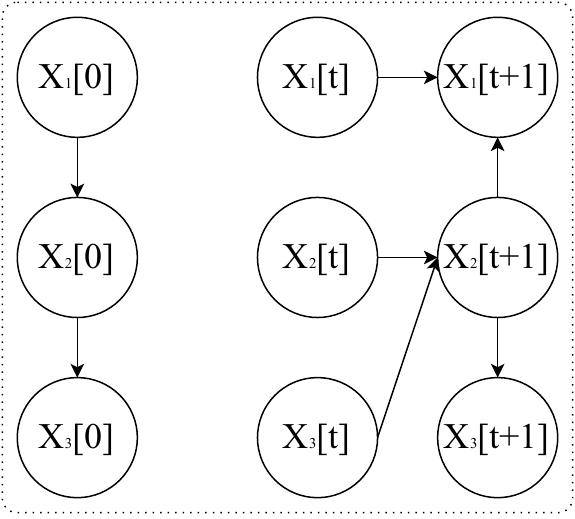}
        \caption{}
        \label{fig:1a}
    \end{subfigure}
    \hfill
    \begin{subfigure}{0.58\textwidth}
        \includegraphics[width=\linewidth]{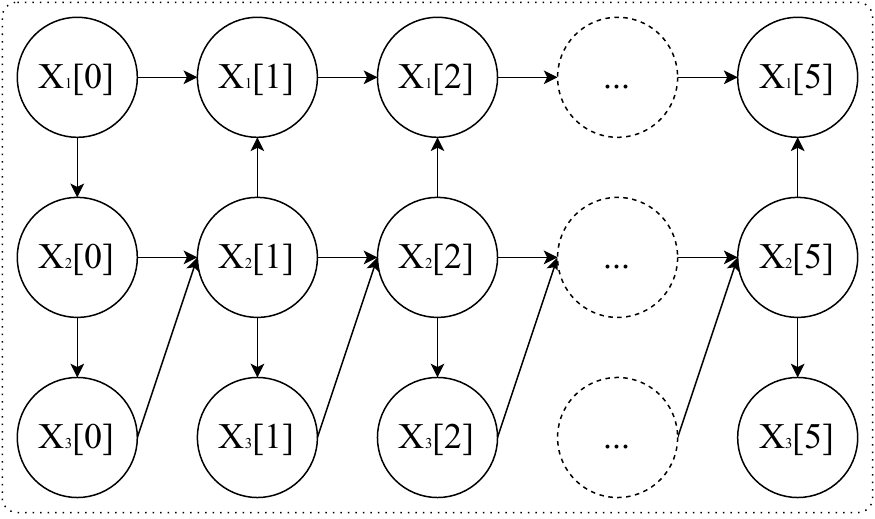}
        \caption{}
        \label{fig:1b}
    \end{subfigure}
    \caption{
        An example of a 2-TBN. 
        (a) The initial model (left) and transition model (right) for a 2-TBN with three variables. 
        (b) The corresponding ``unrolled'' DBN with six time slices.
    }
    \label{fig:dbn}
\end{figure}

Dynamic Bayesian Networks (DBNs)~\cite{murphy2002dynamic} extend static BNs by integrating time information into the original network structure, enabling the processing of time-series data. 
The state changes in a DBN model can be visualized as a series of animation frames, with each frame capturing the current state of the DBN.
These frames, often called time slices, contain a observation of random variables. 
For simplicity, it is commonly assumed that each time slice comprises variables from a consistent set. 
A DBN is a probabilistic graphical model designed to represent sequential systems, determining the probability distribution over $X[t]$, where $X[t] = \{X_1[t], X_2[t], \dots, X_n[t]\}$ represents the $n$ variables observed over discrete time $t$.
DBNs represent a category of graphical models widely recognized as a standard tool for modeling a range of stochastic, time-varying, or non-stationary phenomena. 
While numerous studies employ various forms of DBNs, we focus on a special class of DBNs in this work, namely 2 Time-sliced Bayesian Network (2-TBN) with an example shown in Figure~\ref{fig:dbn}.

A 2-TBN can be defined as a pair of BNs, denoted as $(B_0, B_{\rightarrow})$. 
The BN $B_0 = (G_0, \Theta_0)$ represents the initial joint distribution of the process, expressed as $P(X[t=0])$, while $B_{\rightarrow} = (G_{\rightarrow}, \Theta_{\rightarrow})$ represents the distribution $P(X[t+1] | X[t])$. 
A 2-TBN is a type of DBN that adheres to the first-order Markov property, formulated as $X[t-1] \perp X[t+1] | X[t]$. 
This implies that the variables in the time slice $t + 1$ are independent of those in time slice $t - 1$, given the variables in time slice $t$.
The initial model $B_0$ is a basic static BN, comprising a DAG \(G_0\) with the variables $X[t=0]$ and a set of conditional distributions $P(X_i[t=0] | X_{pa(i)_{G_0}})$, where $X_{pa(i)_{G_0}}$ denotes the parents of the variable $X_i[t=0]$ in $G_0$. 
The transition model $B_{\rightarrow}$ is a two-slice temporal BN, consisting of the DAG $G_{\rightarrow}$ with variables $X[t] \cup X[t+1]$ and a set of conditional distributions $P(X_i[t+1] | X_{pa(i)_{G_{\rightarrow}}})$, where $X_{pa(i)_{G_{\rightarrow}}}$ represents the parents of $X_i[t+1]$ in $G_{\rightarrow}$. 
The distribution of the transition model $B_{\rightarrow}$ is defined as follows: 
\begin{equation}
    P(X[t+1] | X[t]) = \prod_{i=1}^{n} P(X_i[t+1] | X_{pa(i)_{G_{\rightarrow}}})
\end{equation}
The joint probability distribution for a sequence from $t=0$ to $t=T$ is obtained by unrolling the 2-TBN: 
\begin{equation}
    P(X[0:T]) = \prod_{i=1}^{n} P(X_i[t=0] | X_{pa(i)_{G_0}}) \times \prod_{t=0}^{T-1} \prod_{i=1}^{n} P(X_i[t+1] | X_{pa(i)_{G_{\rightarrow}}})
\end{equation}

\subsection{Structure Learning Problem}

Constructing a static BN typically involves two main stages: 
(a) identifying the structure of the DAG, and 
(b) determining the parameter set for all conditional probabilities of the variables, given the states of their parent nodes in the DAG. 
The structure learning problem, as the name suggests, is primarily concerned with identifying the DAG structure of the BN. 
In the case of a DBN represented by $(B_0, B_{\rightarrow})$, structure learning is divided into two components —
structure learning of a straightforward static DAG $G_0$ containing variables $X[t=0]$, 
and structure learning of a time slice DAG $G_{\rightarrow}$ containing variables $X[t] \cup X[t+1]$. 
This graph structure may be identified through expert knowledge, data-driven methodologies, or a blend of both. 
However, manually constructing large-scale BNs or DBNs is often impractical. 
Consequently, this paper primarily focuses on a data-driven approach, where DAGs are inferred from observable data rather than relying on expert experience.

The DBN structure learning problem we discuss in this work is formally defined as follows: 
Given an observable dataset $D = \{d_1, d_2, \ldots, d_m\}$ comprising $m$ data samples, each sample $d_i \in D$ represents a sequence of length $l$ with instances $(x[t=0], x[t=1], \ldots, x[t=l-1])$. 
Every instance contains observed values $\{x_1, x_2, \ldots, x_n\}$ corresponding to the random variables $X = \{X_1, X_2, \ldots, X_n\}$ in a specific time slice. 
The objective is to determine the DAG structures $G_0$ and $G_{\rightarrow}$ of the 2-TBN in a manner that most effectively captures the causal relationships among the random variables within the time-series process.
In practice, a designated structure evaluation metric is typically used to quantify the quality of the derived DAG structure.
As the number of nodes increases, there is a recursive surge in the number of DAGs~\cite{rodionov1992number}. 
This relationship is captured by the following equation:
\begin{equation}
\left|G_{n}\right|=\sum_{i=1}^{n}
(-1)^{i-1}
C_{n}^{i}
2^{i(n-i)}
\left|G_{n-i}\right|
\end{equation}
where $\left|G_{n}\right|$ is the number of DAGs with $n$ nodes, while $C_{n}^{i}$ denotes the combinations of selecting $i$ elements from a total of $n$, and $|G_{0}|$ is defined as $1$.
Furthermore, the structure learning of 2-TBN is classified as an NP-hard problem, as is the case with static BN~\cite{chickering2004large}.

\subsection{Related Work}

Static BN structure learning is an NP-hard problem~\cite{chickering2004large}, typically addressed using approximation methods. 
These methods are categorized into constraint-based, score-based, and hybrid approaches. 
Constraint-based methods, including the PC~\cite{spirtes2000causation}, MMPC~\cite{tsamardinos2003time}, and PC-Stable~\cite{colombo2014order} algorithms, rely on Conditional Independence (CI) tests on observations to discern relationships among variables. 
Within the space of DAGs or Markov Equivalence Classes (MECs), score-based methods use heuristic techniques (Tabu Search (TS)~\cite{bouckaert1995bayesian}, Genetic Algorithm (GA)~\cite{larranaga1996learning}, Greedy Search (GS)~\cite{chickering2002learning}, etc.) and scoring functions (BDeu~\cite{akaike1974new}, BIC~\cite{schwarz1978estimating}, K2~\cite{cooper1992bayesian}, etc.) to steer the search.
Recent developments in score-based methods, such as NOTEARS~\cite{zheng2018dags} and LEAST~\cite{zhu2021efficient}, have reframed BN structure learning as a continuous optimization problem. 
Hybrid methods, exemplified by MMHC~\cite{tsamardinos2006max}, combine constraint-based and score-based strategies, using MMPC to construct the graph skeleton and TS for finalizing the BN. 
As the number of variables grows, traditional methods often slow down significantly and lose accuracy. 
To address this, fGES~\cite{ramsey2017million}, building on GES~\cite{chickering2002learning}, has demonstrated the ability to manage large-scale problems by narrowing the search space and facilitating parallel search. 
Additionally, the Partition-Estimation-Fusion (PEF)~\cite{gu2020learning} strategy applies a divide-and-conquer approach, significantly improving the handling of large-scale problems.

Current techniques for learning DBNs structure are largely adaptations of those used in BN structure learning.
To illustrate, DBN structure learning can be divided into two independent stages: learning the initial graph $G_0$ as a static BN structure with a dataset for $X[t = 0]$, and learning the transition graph $G_{\rightarrow}$ with another ``static'' dataset for all the transitions $X[t] \cup X[t + 1]$. 
Building on this, \cite{friedman1998learning} has seamlessly integrated the use of GS in BN structure learning into DBN structure learning. 
The DMMHC~\cite{trabelsi2013dynamic} algorithm, specifically designed for DBN structure learning, evolves from the MMHC algorithm used in static one. 
Additionally, numerous studies explore the application of bayesian optimization and evolutionary computation to address this challenge~\cite{gao2007learning,lou2015structure,wang2006learning}. 
When dealing with partially observed systems (incomplete data), structure learning becomes computationally demanding. 
One effective solution is the Structural Expectation-Maximization (SEM)~\cite{tsamardinos2006max} algorithm. 
While these methods have been validated on small benchmark models, they tend to become exceedingly complex as the number of variables increases, reflecting the inherent limitations of current structure learning methodologies.

\section{Proposed Method}\label{sec:pm}

Due to the inherent limitations of current structure learning methodologies, algorithms tend to become increasingly complex and time-consuming with the addition of more variables. 
The divide-and-conquer approach is a widely employed strategy for addressing large-scale problems in algorithm design. 
This strategy provides significant advantages, particularly when existing structure learning algorithms exhibit time complexity beyond linear, enhancing scalability. 
The Partition-Estimation-Fusion (PEF)~\cite{gu2020learning} strategy, based on the divide-and-conquer principle~\cite{LiuT019,TangLYY21,LiuTY22}, has been developed for large-scale structure learning in static Bayesian Networks (BNs). 
Works such as \cite{friedman1998learning,trabelsi2013dynamic} approach the problem of Dynamic Bayesian Network (DBN) structure learning by transforming it into two separate static BN structure learning problems. 
We believe that the PEF strategy can be effectively adapted for the 2 Time-sliced Bayesian Network (2-TBN) structure learning, yielding relevant results.

In essence, the structure learning problem of a 2-TBN can be divided into two distinct ``static'' tasks: 
\begin{itemize}
    \item Extracting all observation data in time slices $t = 0$ from the time series into a dataset for $X[t = 0]$ to learn the initial graph $G_0$, and
    \item Combining observation data from adjacent time slices into a single data point, forming another ``static'' dataset for $X[t] \cup X[t + 1]$ to learn the transition graph $G_{\rightarrow}$.  
\end{itemize}
Following this transformation, the PEF strategy can be employed to learn $G_0$ and $G_{\rightarrow}$ separately. 
The PEF strategy comprises the following three steps:
\begin{enumerate}
    \item Partition - The nodes (variables) are divided into clusters using a Modified Hierarchical Clustering (MHC) algorithm.
    \item Estimation - Applying an existing structure learning method to estimate a subgraph for each cluster.
    \item Fusion - Merging estimated subgraphs into a comprehensive DAG encompassing all nodes. 
\end{enumerate}
Besides introducing the PEF strategy following the transformation of 2-TBN structure learning, this work also includes enhancements to the fusion phase of PEF to fully leverage the prior knowledge of 2-TBN.
In the remainder of this section, we will give the details of the partition and estimation phases in Section~\ref{subsec:pae}, and the enhanced fusion techniques for 2-TBN in Section~\ref{subsec:fusion}.

\subsection{Partition and Estimation}\label{subsec:pae}

Consider a static observation data matrix $D_{static}$ comprising $m$ samples. 
Each node in the set $V = \{V_1, V_2, \dots, V_n\}$ corresponds to a data column $X_j \in {R}^m$ for $j = 1, 2, \dots, n$. 
The Partition step (P-step) in the PEF strategy involves grouping nodes into clusters. 
This step results in $p$ clusters, labeled as $C_i$ for $i = 1, 2, \dots, p$, using a MHC method with average linkage. 
This method autonomously determines the number of clusters $p$ as follows.
The PEF strategy considers that the minimum size for clusters should be $0.05n$, referring to these as ``big clusters''. 
For each iteration $h = 0, 1, \dots, n-1$, $C_h$ represents the clusters formed during the $h^{th}$ iteration of bottom-up hierarchical clustering. 
Specifically, $C_0 = \{\{V_1\}, \{V_2\}, \dots, \{V_n\}\}$ consists of $n$ singleton clusters, and $C_{n-1} = \{\{V_1, V_2, \dots, V_n\}\}$ denotes a single cluster comprising all $n$ nodes. 
The number of big clusters in $C_i$ is denoted by $p_i$. 
PEF determines a specific $p$ based on the equation:
\begin{equation}\label{eq:k}
    p = \min\{p_{max},\max_{0 \leq i \leq n-1}p_{i}\}
\end{equation}
where $p_{max} \leq 20$ is the user-defined maximum number of big clusters.
After determining $p$, the highest level $l$ on the dendrogram containing $p$ big clusters is defined by:
\begin{equation}\label{eq:l}
    l = \arg\max_{0 \leq i \leq n-1}\{i : p_i = p\}
\end{equation}
Clusters at level $C_l$ are reordered in descending order of their sizes, such that $S_1 \geq S_2 \geq \dots \geq S_{n-l}$, where $S_i = |C_i|$. 
The first $p$ clusters are identified as the primary big clusters. 
MHC then integrates remaining small clusters into these big clusters by continuously merging the closest pairs.
In PEF, the distance between two nodes $V_i$ and $V_j$ is defined as:
\begin{equation}\label{eq:dist}
    d(i, j) = 1 - \left| r_{ij} \right| \in [0, 1]
\end{equation}
where $r_{ij} = {cor}(X_{i}, X_{j})$ denotes the correlation between $X_i$ and $X_j$ for $i, j = 1, 2, \dots, n$. This correlation is calculated using the covariance $cov(X_i, X_j)$, standard deviations $\sigma_{X_i}, \sigma_{X_j}$, and the formula:
\begin{equation}\label{eq:cor}
    {cor}(X_{i}, X_{j}) = \frac{cov(X_{i}, X_{j})}{\sigma_{X_i} \sigma_{X_j}}
\end{equation}
The pseudocode for the MHC algorithm is provided in Algorithm \ref{alg:mhc}. 
It starts with the computation of the dissimilarity matrix $D = (d(i, j))_{n \times n}$, as defined by Equation~\ref{eq:dist}.

\begin{algorithm}[t]
    \caption{Modified Hierarchical Clustering (MHC)}
    \label{alg:mhc}
    \begin{algorithmic}[1] 
        \STATE Perform hierarchical clustering using the dissimilarity matrix $D = (d(i, j))_{n \times n}$.
        \STATE Construct the dendrogram $T_{D}$ from the hierarchical clustering.
        \STATE Determine $p$ using Equation~\ref{eq:k} and $l$ using Equation~\ref{eq:l}.
        \STATE Reorder clusters in $C \leftarrow C_l$ such that $S_{1} \geq \dots \geq S_{n-l}$.
        \WHILE{$|C| > p$}
            \STATE $(i^{*},j^{*}) \leftarrow {argmin}_{(i,j)}\{d(C_{i},C_{j}): i<j \land j > p\}$.
            \STATE $C_{i^{*}} \leftarrow C_{i^{*}} \cup C_{j^{*}}, C \leftarrow C \backslash \{C_{j^{*}}\}$.
        \ENDWHILE
        \STATE \textbf{return} the final clusters $C = \{C_{1}, C_{2}, \dots, C_{p}\}$.
    \end{algorithmic}
\end{algorithm}

During the Estimation step (E-step) of the PEF process, the structure of each subgraph is determined individually. 
This step functions as a black box within the PEF, allowing users to apply any BN structure learning algorithm for estimating subgraphs without requiring detailed technical knowledge. 
Usually, this step results in $p$ Partial DAGs (PDAGs). 
It is important to note that both DAGs and complete PDAGs (CPDAGs) are included within the category of PDAGs. 
If the time complexity of a chosen structure learning technique exceeds $O(n)$, the time to learn small subgraphs in the E-step is significantly reduced compared to estimating a complete DAG. 
At the same time, E-step is easy to parallelize.
Assuming that nodes have been divided into $p$ clusters $C_1, C_2, \ldots, C_p$ during the P-step, and the time to learn a PDAG on $C_i$ is $t_i$, executing the learning process of $p$ subgraphs in parallel across $p$ cores can minimize the duration of the E-step to $\max\{t_i: i = 1, 2, \ldots, p\}$. 
This time is typically influenced by the size of the largest cluster.

\subsection{Enhanced Fusion for 2-TBN}\label{subsec:fusion}

During the Fusion step (F-step) of the PEF process, a hybrid method is applied to derive the full DAG structure from the subgraphs estimated in the E-step. 
The F-step involves two phases. 
Initially, PEF creates a candidate edge set $A$ to narrow down the search space. 
It then identifies a subset $A^{*}$, consisting of potential edges between subgraphs, through statistical tests. 
As a result, the candidate edge set $A$ includes $A^{*}$ along with all edges identified in each subgraph during the E-step. 
Subsequently, PEF optimizes the DAG structure by iteratively refining edges within set $A$ using a modified Bayesian Information Criterion (BIC) score. 
The final outcome of the F-step is a DAG.
Building on the original F-step, this study incorporates specific prior knowledge of the 2-TBN to enhance the F-step. 
This prior knowledge is primarily applied in learning the transition graph $G_{\rightarrow}$: 
\begin{itemize}
    \item Interactions among variables within previous time slice $t$ are not considered, hence there are no connections between nodes belonging to $X[t]$;
    \item Future observations cannot influence past variables, therefore edges from $X[t+1]$ to $X[t]$ are excluded. 
\end{itemize}

\subsubsection{Finding Candidate Edge Set}

For subgraphs denoted as $G_m$ ($m = 1, 2, \dots, p$), let $z(i) \in \{1, 2, \dots, p\}$ represent the cluster label of node $V_i$. 
Typically, these subgraphs $G_m$ are represented as PDAGs. 
Neighbors of node $V_i$ in subgraph $G_{z(i)}$ are defined by $N_i(z(i)) = \{V_j \in G_{z(i)} : V_j \rightarrow V_i \in G_{z(i)} \lor V_i - V_j \in G_{z(i)}\}$, where $V_j \rightarrow V_i$ indicates a directed edge, and $V_i - V_j$ an undirected one. 
The correlation ${\widetilde{\rho}}_{ij}$ between the residuals $R_i$ and $R_j$, after projecting $X_i$ onto its neighbors $N_i(z(i))$ in $G_{z(i)}$, is used to filter unlikely edges between subgraphs. 
Specifically, an initial candidate set ${\widetilde{A}}^{*}$ includes pairs $(V_i, V_j)$ where $z(i) \neq z(j)$ and the hypothesis ${\widetilde{\rho}}_{ij} = 0$ is rejected at a significance level $\alpha$ using a z-test with Fisher transformation on the correlation coefficient ${\widetilde{\rho}}_{ij}$. 
Subsequently, a sequential method refines ${\widetilde{A}}^{*}$ to determine the final candidate edge set $A^{*}$ between subgraphs. 
Each node pair $(V_i, V_j)$ in ${\widetilde{A}}^{*}$ undergoes a conditional independence test considering the union of their updated neighbors, $N_i(z(i)) \cup N_j(z(j)) \cup P_{ij}$, where $P_{ij}$ is defined as the set of neighbors of $V_i$ or $V_j$ in the current candidate set $A^{*}$ between subgraphs. 
Let $G$ be the PDAG composed of disconnected subgraphs derived from the E-step, and $SK(G)$ represent the edge set in the skeleton of $G$, that is, $SK(G) = \{(V_i, V_j): V_i - V_j \in G \lor V_i \rightarrow V_j \in G\}$. 
The candidate edge set $A$ is generated by appending $SK(G)$ to $A^{*}$. 
The edges in the final output DAG of the method will be a subset of $A$. 
The pseudocode for finding the candidate edge set $A$ is outlined in Algorithm \ref{alg:candidate}, where $I_{p}(X_{i}; X_{j} | Z)$ denotes an independence test. In implementation, PEF sorts the node pairs in ${\widetilde{A}}^{*}$ in ascending order of their p-values in testing against ${\widetilde{\rho}}_{ij} = 0$.

\begin{algorithm}[t]
    \caption{Find Candidate Edge Set $A$}
    \label{alg:candidate}
    \begin{algorithmic}[1] 
        \STATE Input data matrix $D_{static}$ and estimated subgraphs $G_{1}, G_{2}, \dots, G_{p}$.
        \STATE Set ${\widetilde{A}}^{*} = \emptyset$.
        \FOR{all pairs $(V_i, V_j)$ such that $z(i) \neq z(j)$} \label{alg:line:pair}
            \IF{${\widetilde{\rho}}_{ij} = 0$ is rejected at level $\alpha$}
                \STATE ${\widetilde{A}}^{*} \leftarrow {\widetilde{A}}^{*} \cup (V_i, V_j)$.
            \ENDIF
        \ENDFOR
        \STATE Set $A^{*} = \emptyset$.
        \FOR{all $(V_i, V_j) \in {\widetilde{A}}^{*}$}
            \STATE Let $Z = N_i(z(i)) \cup N_j(z(j)) \cup P_{ij}$.
            \IF{$I_p(X_i; X_j | Z)$ is rejected at level $\alpha$}
                \STATE $A^{*} \leftarrow A^{*} \cup (V_i, V_j)$.
            \ENDIF
        \ENDFOR
        \STATE \textbf{return} $A = A^{*} \cup SK(G)$.\label{alg:line:rnt}
    \end{algorithmic}
\end{algorithm}

\subsubsection{Learning Full DAG Structure}

In the final phase of the F-step, the PEF method determines the existence and direction of each edge for node pairs $(V_i, V_j) \in A$. 
This is accomplished by sequentially minimizing a modified BIC score, known as the Risk Inflation Criterion (RIC), over the candidate edge set. The RIC score, defined as:
\begin{equation}\label{eq:ric}
    RIC(G) = -2l(G) + \lambda d(G)
\end{equation}
which includes two components: 
a log-likelihood component $l(G)$, evaluating how well graph $G$ fits the data; and 
a regularization term $d(G)$, weighted by $\lambda = 2\log n$, to encourage sparsity. 
The PEF employs this score especially when dealing with a large number of nodes ($n > \sqrt{m}$). 
Conversely, for $n \leq \sqrt{m}$, the standard BIC score with $\lambda = \log m$ is used.
For every pair $(V_i, V_j) \in A$, PEF evaluates three models: 
\begin{itemize}
    \item $M_0$ - Absence of an edge between $V_i$ and $V_j$.
    \item $M_1$ - $V_i$ as a parent of $V_j$.
    \item $M_2$ - $V_j$ as a parent of $V_i$. 
\end{itemize}
During this evaluation, all other edges in $G$ remain fixed. 
An edge is introduced between $V_i$ and $V_j$ if $\max\{RIC(M_1), RIC(M_2)\} < RIC(M_0)$. 
Subsequently, PEF decides the direction of the edge, ensuring acyclicity or opting for the model with the lower RIC score based on a predefined tie-breaker. 
Algorithm \ref{alg:fuse} illustrates the F-step, which iteratively processes $A$ until no further changes in the structure of $G$ are observed. 
Upon completion of the F-step, PEF returns the finalized DAG.

\begin{algorithm}[t]
    \caption{Fuse Subgraphs}
    \label{alg:fuse}
    \begin{algorithmic}[1] 
        \STATE Input data matrix $D_{static}$ and estimated subgraphs $G_{1}, G_{2}, \dots, G_{p}$.
        \STATE Run Algorithm \ref{alg:candidate} to generate candidate edge set $A$.
        \STATE Initialize $G$ as the PDAG comprising $G_1, G_2, \dots, G_p$.
        \FOR{each pair $(V_i, V_j) \in A$} \label{alg:line:loop_in}
            \IF{$V_i, V_j$ are adjacent in $G$}
                \STATE Remove the edge from $G$.
            \ENDIF
            \IF{$I_p(X_i; X_j | N_i(G) \cup N_j(G))$}
                \STATE Remove $(V_i, V_j)$ from $A$.
            \ELSE \label{alg:line:c1}
                \STATE Set ${RIC}_{\max} = \max\{RIC(M_1), RIC(M_2)\}$.
                \IF{${RIC}_{\max} < RIC(M_0)$}
                    \IF{adding edge $V_i \rightarrow V_j$ induces a cycle}
                        \STATE Add edge $V_j \rightarrow V_i$ to $G$.
                    \ELSIF{adding edge $V_j \rightarrow V_i$ induces a cycle}
                        \STATE Add edge $V_i \rightarrow V_j$ to $G$.
                    \ELSE 
                        \STATE Choose the direction leading to a smaller $RIC$.
                    \ENDIF
                \ENDIF
            \ENDIF \label{alg:line:c2}
        \ENDFOR \label{alg:line:loop_out}
    \STATE Repeat lines \ref{alg:line:loop_in} to \ref{alg:line:loop_out} until the structure of $G$ remains unchanged. 
    \STATE \textbf{return} $G$.
    \end{algorithmic}
\end{algorithm}

\subsubsection{Utilization of Prior Knowledge}

For the structure learning of the initial graph $G_0$, no prior knowledge is available. 
Therefore, the F-step as outlined in Algorithms~\ref{alg:candidate} and \ref{alg:fuse} can be directly applied. 
Conversely, for the structure learning of the transition graph $G_{\rightarrow}$, the previously mentioned prior knowledge can be utilized to enhance the F-step. 
The improvements are as follows: 
In line~\ref{alg:line:pair} of Algorithm~\ref{alg:candidate}, pairs $(V_i, V_j)$ where both $V_i$ and $V_j$ are in $X[t]$ should be skipped. 
In line~\ref{alg:line:rnt} of Algorithm~\ref{alg:candidate}, pairs $(V_i, V_j) \in SK(G)$ should be removed if both $V_i$ and $V_j$ are in $X[t]$.
From line~\ref{alg:line:c1} to \ref{alg:line:c2} in Algorithm~\ref{alg:fuse}, do not consider directed edge $V_j \rightarrow V_i$ if $V_j \in X[t+1]$ and $V_i \in X[t]$, and vice versa.

\section{Experiments}\label{sec:exp}

This section experimentally validates the effectiveness of the divide-and-conquer strategy for Dynamic Bayesian Network (DBN) structure learning proposed in this work. 
Initially, Section \ref{subsec:data} will detail the sources of our experimental data, followed by a discussion of the experimental setup in Section \ref{subsec:setup}. 
Subsequently, Section \ref{subsec:res} will present a comparative analysis of the divide-and-conquer strategy against baselines, thereby illustrating its overall efficacy.

\subsection{Data Generation}\label{subsec:data}

To overcome the limitations of a small variable count in existing public datasets, we have adapted a method from previous research \cite{gu2020learning,trabelsi2013benchmarking} to create diverse, large-scale 2 Time-sliced Bayesian Network (2-TBN) datasets that meet our experimental needs. 
Our data generation approach, slightly modified from the existing method, is described as follows:
\begin{enumerate}
    \item Duplicate an existing network structure a specified number of times. \label{item:1}
    \item While ensuring the final network remains a Directed Acyclic Graph (DAG), interconnect the duplicates by adding $10\%$ of random edges between them. \label{item:2}
    \item Apply step~\ref{item:1} and \ref{item:2} to generate an initial graph $G_0$.
    \item Apply step~\ref{item:1} and \ref{item:2} to generate the nodes and intra-connections of $X[t+1]$ in the transition graph $G_{\rightarrow}$.
    \item Tile the nodes of $X[t+1]$ in $G_{\rightarrow}$ to nodes of $X[t]$ (excluding intra-connections) and randomly introduce $20\%$ of inter-connections from $X[t]$ to $X[t+1]$.
    \item Simulate numerous observational data records by $G_0$ and $G_{\rightarrow}$.
    \item Normalize the observed data to ensure uniform mean and standard deviation across all data columns.
\end{enumerate}

The structures of all networks were obtained from the repository of bnlearn R package~\cite{scutari2010learning}. 
Within this repository, we selected 10 well-known static networks: Alarm, Asia, Cancer, Child, Earthquake, Hailfinder, Healthcare, Mildew, Pigs, and Survey. 
For each static network, we generated a corresponding large-scale 2-TBN with over 1,000 nodes and sampled 1,000 time series sequences of length 6. 
Consequently, for each problem instance, we have 1,000 $X[t=0]$ samples for $G_0$ structure learning and 5,000 $X[t] \cup X[t+1]$ samples for $G_{\rightarrow}$ structure learning. 
The details of the number of nodes and edges for each 2-TBN problem instance are presented in Table~\ref{tab:overview}.

\begin{table}[t]
    \caption{Number of nodes (variables) and edges for each problem instance.}\label{tab:overview}
    \begin{center}
    \begin{tabular}{@{}cccc@{}}
        \toprule
        \textbf{Instance} & \textbf{\#Nodes} & \textbf{\#Edges of $G_0$} & \textbf{\#Edges of $G_{\rightarrow}$} \\ \midrule
        Alarm             & 1036             & 1417                   & 1701                               \\
        Asia              & 1000             & 1100                   & 1320                               \\
        Cancer            & 1000             & 880                    & 1056                               \\
        Child             & 1000             & 1375                   & 1650                               \\
        Earthquake        & 1000             & 880                    & 1056                               \\
        Hailfinder        & 1008             & 1307                   & 1569                               \\
        Healthcare        & 1001             & 1416                   & 1700                               \\
        Mildew            & 1015             & 1468                   & 1762                               \\
        Pigs              & 1323             & 1954                   & 2345                               \\
        Survey            & 1002             & 1103                   & 1324                               \\ \bottomrule
        \end{tabular}\end{center}
    \end{table}    

\subsection{Experimental Methodology}\label{subsec:setup}

Since our divide-and-conquer strategy is built upon existing Bayesian Network (BN) structure learning algorithms (i.e., base algorithms), we selected using the PC-Stable~\cite{colombo2014order} algorithm as the base algorithm specifically. 
This choice is due to PC-Stable's representation as a well-performing and stable algorithm among current offerings. 
For our experiments, we utilized the PC-Stable implementation from the TETRAD causal discovery toolbox~\cite{ramsey2018tetrad}. 
In our experiments, we conducted runs of the PC-Stable algorithm both with and without (as baselines) using our divide-and-conquer strategy. 
Given that the benefits of the divide-and-conquer approach in static BN structure learning have already been established in \cite{gu2020learning}, our primary focus is on enhancements related to the transition model $G_{\rightarrow}$. 
Therefore, our comparative analysis specifically targets the experimental results associated with the transition model $G_{\rightarrow}$.

In terms of performance evaluation, given our knowledge of the true graph structures in the generated problem instances, we employ edge classification-based metrics. 
This involves classifying the relationship between nodes $A$ and $B$ as either $A \leftarrow B$, $A \rightarrow B$, or no edge, and accordingly, we use the F1 score to assess edge classification accuracy. 
Notably, many BN structure learning algorithms, including PC-Stable, estimate graphs as Markov Equivalence Classes (MECs) of DAGs rather than as DAGs themselves. 
This results in graphs with undirected edges ($A-B$). 
Following the approach in \cite{ramsey2017million}, we evaluate performance using two metrics: F1 Adjacent (disregarding direction) and F1 Arrowhead (considering direction).
We also measure the running time (physical time) of the algorithms in seconds. 
Unless stated otherwise, all experiments were conducted on a Linux server equipped with double Intel(R) Xeon(R) Gold 6336Y CPU @ 2.40GHz, 96 cores, and 768GB of RAM, running Ubuntu 22.04.2 LTS. Due to computational resource constraints, we imposed a one-day runtime limit for each problem instance.

\begin{table}[t]
    \caption{
        The results (mean value ± standard deviation) for testing problem instances are presented in terms of F1 Adjacency (F1 Adj, $\times 10^2$), F1 Arrowhead (F1 Arr, $\times 10^2$), and Runtime (in seconds). 
        In the comparison, metrics that significantly outperform others, indicated by higher mean values and passing significance tests (Wilcoxon signed-rank test at a $99\%$ confidence level), are highlighted in \textbf{bold}.
    }\label{tab:results}
    \begin{center}
        \resizebox{1.00\columnwidth}{!}{
        \begin{tabular}{@{}ccccccc@{}}
            \toprule
            \multirow{2}{*}{Instance} & \multicolumn{3}{c}{PC-Stable}                       & \multicolumn{3}{c}{Our Method}                                         \\ \cmidrule(l){2-7} 
                                      & F1 Adj             & F1 Arr    & Runtime          & F1 Adj              & F1 Arr              & Runtime                  \\ \midrule
            Alarm                     & 39.34±0.53          & 31.15±0.53 & 27411.87±2379.08 & \textbf{62.92±2.25}  & \textbf{57.90±2.25}  & \textbf{322.83±247.25}   \\
            Asia                      & 35.17±0.60          & 21.66±0.60 & 13564.81±930.42  & \textbf{68.62±0.65}  & \textbf{61.23±0.65}  & \textbf{64.85±24.40}     \\
            Cancer                    & 23.54±0.40          & 13.55±0.40 & 16661.04±3284.75 & \textbf{57.64±0.68}  & \textbf{47.79±0.68}  & \textbf{70.19±21.44}     \\
            Child                     & 46.36±0.71          & 37.45±0.71 & 24357.04±2921.17 & \textbf{74.08±1.09}  & \textbf{66.94±1.09}  & \textbf{386.68±537.49}   \\
            Earthquake                & 23.57±0.33          & 13.81±0.33 & 16139.89±1984.97 & \textbf{57.73±0.91}  & \textbf{47.67±0.91}  & \textbf{51.94±7.48}      \\
            Hailfinder                & 35.14±0.27          & 25.14±0.27 & 29948.92±1397.61 & \textbf{49.27±22.43} & \textbf{44.63±22.43} & \textbf{2446.96±3079.47} \\
            Healthcare                & 45.64±0.71          & 41.09±0.71 & 14277.13±1081.79 & \textbf{76.06±1.19}  & \textbf{69.37±1.19}  & \textbf{289.80±619.64}   \\
            Mildew                    & \textbf{31.14±1.01} & 17.43±1.01 & 31279.77±2137.75 & 27.47±14.73          & \textbf{23.01±14.73} & \textbf{3632.29±3243.75} \\
            Pigs                      & 0.00±0.00           & 0.00±0.00  & 86400.00±0.00    & \textbf{6.46±1.90}   & \textbf{5.38±1.90}   & \textbf{9822.01±1595.96} \\
            Survey                    & 33.37±0.66          & 28.68±0.66 & 10799.27±1011.99 & \textbf{66.26±0.93}  & \textbf{61.16±0.93}  & \textbf{105.03±157.52}   \\ \bottomrule
            \end{tabular}}\end{center}
    \end{table}   

\subsection{Results and Discussion}\label{subsec:res} 

In this part, we assess the performance of our divide-and-conquer strategy in comparison to its base algorithm, PC-Stable. 
To ensure robustness and reliability of our experimental results~\cite{LiuTL020}, 10 unique problem instances for each base network were generated using varied random seeds. 
The outcomes are reported as “mean value ± standard deviation”.
For each comparative experimental set, a Wilcoxon signed-rank test at a $99\%$ confidence level was conducted to evaluate the presence of significant differences in the results.
The experimental results are presented in Table~\ref{tab:results}.
The F1 score is marked as zero if the algorithm fails to yield a reasonable result within one day.
Metrics highlighted in \textbf{bold} underscore instances where our strategy demonstrates a significantly superior performance compared to the base algorithms, marked by higher mean values and concurrent success in significance tests. 
As indicated by the results in Table~\ref{tab:results}, our divide-and-conquer strategy generally exhibits remarkable effectiveness. 
It consistently surpasses the base algorithm PC-Stable across most metrics. 
Notably, in the one instance where our method's F1 adjacent score does not significantly outperform the baselines, the discrepancy remains minimal, while the runtime is notably shorter. 
Moreover, our proposed approach on average improves two accuracy metrics by $74.45\%$ and $110.94\%$, respectively, while reducing runtime by an average of $93.65\%$.
Hence, under this experimental conditions, our divide-and-conquer strategy has demonstrated its efficacy in the structure learning of 2-TBN.
It is worth mentioning that in the E-step of the divide-and-conquer strategy, any existing algorithm can be used to solve the sub-problem, making the method highly portable.
We also conducted experiments based on other base algorithms and finally obtained similar experimental results.

\section{Conclusion}\label{sec:con}

In this paper, we address the challenge of large-scale Dynamic Bayesian Network (DBN) structure learning by implementing a divide-and-conquer strategy, originally developed for static Bayesian Network (BN) structure learning. 
Our approach enhances the existing Partition-Estimation-Fusion (PEF) strategy by fully leveraging the prior knowledge of 2-Time-sliced Bayesian Networks (2-TBN), a specific class of DBNs, to improve the learning of the transition model structure in 2-TBNs. 
Our experimental findings confirm the effectiveness of our approach.
Looking ahead, future research will explore the extension of the divide-and-conquer strategy to various types of DBNs.


%
%
%
\bibliographystyle{splncs04}
\bibliography{dbn_pef_ref}

\end{document}